\documentclass[letterpaper, 10 pt, conference]{ieeeconf}  

\IEEEoverridecommandlockouts                              

\usepackage{algorithm}
\usepackage{algorithmic}
\usepackage{amssymb}
\usepackage{subfigure}
\usepackage{booktabs}
\usepackage{multirow}

\usepackage{graphicx}
\usepackage{cite}
\usepackage{soul}
\usepackage{amsmath}
\usepackage{amssymb}
\usepackage{threeparttable}

\usepackage{color}
\definecolor{red}{rgb}{0.9,0.1,0}
\definecolor{gray}{rgb}{0.8,0.8,0.8}
\definecolor{blue}{rgb}{0.4,0.4,0.9}
\definecolor{green}{rgb}{0, 0.4, 0}
\definecolor{orange}{rgb}{1, 0.5, 0}
\definecolor{slateblue}{rgb}{0.7,0.35,0.9}
\definecolor{mahogany}{rgb}{0.75, 0.25, 0.0}
\definecolor{purple}{rgb}{0.6, 0, 0.6}
\definecolor{goldenrod}{rgb}{0.85, 0.65, 0.13}
\newboolean{revising}
\setboolean{revising}{false}
\ifthenelse{\boolean{revising}}
{
    \newcommand{\fong}[1]{\textcolor{magenta}{#1}}

} {
    
    \newcommand{\fong}[1]{#1}

}

\title{\LARGE \bf
The Study of Highway for Lifelong Multi-Agent Path Finding
}

\author{
Ming-Feng Li\ 
\and
Min Sun\ 
\thanks{Authors are with
National Tsing Hua University, Hsinchu City 300, Taiwan (e-mail: justin.mingfeng.li@gmail.com; sunmin@ee.nthu.edu.tw).}
}

\begin{document}

\maketitle
\thispagestyle{empty}
\pagestyle{empty}


\begin{abstract}
    In modern fulfillment warehouses, agents traverse the map to complete endless tasks that arrive on the fly, which is formulated as a lifelong Multi-Agent Path Finding (lifelong MAPF) problem.
    The goal of tackling this challenging problem is to find the path for each agent in a finite runtime while maximizing the throughput.
    However, existing methods encounter exponential growth of runtime and undesirable phenomena of deadlocks and rerouting as the map size or agent density grows.
    To address these challenges in lifelong MAPF, we explore the idea of highways mainly studied for one-shot MAPF (i.e., finding paths at once beforehand), which reduces the complexity of the problem by encouraging agents to move in the same direction.
    We utilize two methods to incorporate the highway idea into the lifelong MAPF framework and discuss the properties that minimize the existing problems of deadlocks and rerouting.
    The experimental results demonstrate that the runtime is considerably reduced and the decay of throughput is gradually insignificant as the map size enlarges under the settings of the highway.
    Furthermore, when the density of agents increases, the phenomena of deadlocks and rerouting are significantly reduced by leveraging the highway.
\end{abstract}

\section{Introduction}
%

Multi-Agent Path Finding (MAPF) is the problem of planning collaborative paths for a team of agents while avoiding collisions. MAPF has been widely used in applications like video games~\cite{ma2017feasibility}, traffic management~\cite{ho2019multi}, and delivery policies~\cite{ma2017lifelong}.
Several MAPF solvers has been proposed in the past years, such as CA*~\cite{silver2005cooperative}, CBS~\cite{sharon2015conflict}, ECBS~\cite{barer2014suboptimal}, which consider the path planning problem in a one-shot manner, assuming each agent has only one pair of start and goal locations. Generally, these methods are not suitable for applications in online scenarios \fong{without reasonable modification.}

Considering applications in the real world, especially for robots in fulfillment warehouses~\cite{wurman2008coordinating}, instead of having only one pair of start and goal locations, new goals are assigned on the fly. This scenario is referred to as lifelong MAPF. Although current solutions for lifelong MAPF \cite{vcap2015complete, wan2018lifelong, nguyen2019generalized, grenouilleau2019multi} can handle a sequence of goals, they present poor computational efficiency or low throughput (i.e, finished tasks per timestep). 
Therefore, Rolling-Horizon Collision Resolution (RHCR)~\cite{li2021lifelong} incorporates the idea of windowed MAPF~\cite{silver2005cooperative} into the lifelong MAPF framework, separating a lifelong MAPF problem into a sequence of one-shot MAPF problems, which gains computational efficiency without sacrificing throughput.
Nevertheless, the approach still encounters undesirable phenomena of deadlocks and rerouting~\cite{silver2005cooperative, li2021lifelong}.

\begin{figure}[t]
    \centering   \includegraphics[width=0.87\columnwidth]{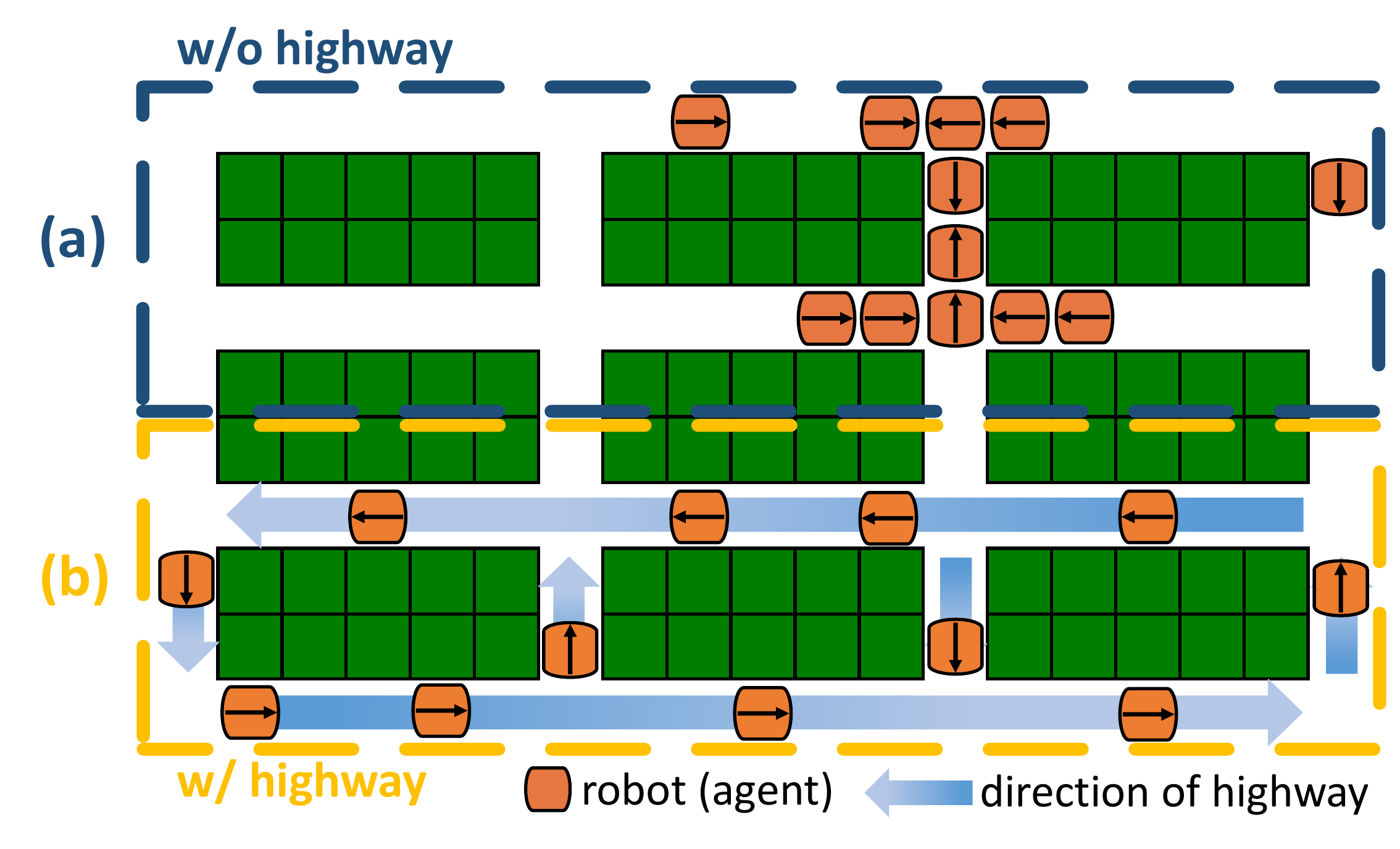}
    \vspace{-1mm}
    \caption{Two scenarios in a warehouse map. Panel (a) shows a map without the highway. Panel (b) shows a map with the highway to solve the situation of crowded agents contributing to deadlock and rerouting in panel (a). The highway direction is represented as blue arrows.
    }
    \label{fig:teaser}
    \vspace{-4mm}
\end{figure}

In order to solve such challenges in lifelong MAPF, we study the methods of \emph{highway} typically utilized in one-shot MAPF~\cite{jansen2008direction, wang2008fast, cohen2015feasibility}.
First, we leverage a directed map~\cite{jansen2008direction, wang2008fast} by enforcing agents to move along specific directions. This simple yet effective strategy decreases the complexity of the problem, which translates into higher efficiency. Additionally, instead of enforcing a movement direction, we leverage ~\cite{cohen2015feasibility} to calculate heuristic values that penalize movements against the highway direction during planning. In short, both strategies encourage agents to move in the same direction and thus effectively avoid face-to-face conflicts.
Subsequently, we further discuss the properties after incorporating \emph{highway} into the lifelong MAPF framework, which can effectively minimize the undesirable phenomena of deadlocks and rerouting, as shown in Fig.~\ref{fig:teaser}.

Through several experiments, we present the benefits of utilizing \emph{highway} in the lifelong MAPF framework and the trade-off between runtime and throughput. The experimental results show that the runtime can be accelerated dozens of times with less than 10\% throughput decay in warehouse-like maps with more than 50\% obstacles.
Furthermore, when the density of agents is high, the throughput even increases after utilizing \emph{highway}, because the severe impacts of deadlocks and rerouting are significantly reduced by \emph{highway}.

\section{Background and Related Work}
\subsection{One-shot MAPF}
MAPF is an NP-hard problem~\cite{surynek2010optimization, yu2013structure} and is typically solved in one-shot and offline, where each agent has exactly one goal known beforehand and paths of agents are found at once, respectively. The solution of one-shot MAPF solvers is typically evaluated by \emph{sum-of-costs} (the sum of the arrival times of each agent) or \emph{makespan} (the time span for all agents to reach their goals).

There are various one-shot MAPF methods~\cite{felner2017search}, including compilation-based solvers~\cite{surynek2019unifying, lam2022branch}, rule-based solvers~\cite{luna2011push, de2013push}, A*-based solvers~\cite{goldenberg2014enhanced, wagner2015subdimensional}, and prioritized planning~\cite{okumura2022priority}, etc. Especially, the search-based solvers~\cite{sharon2013increasing, barer2014suboptimal} and their variants are common. Most of the search-based solvers are two-level MAPF solvers. The low-level solver plans the paths for each agent and the high-level solver handles the conflicts of these paths to guarantee no collisions.
A*~\cite{hart1968formal} is a typical algorithm as the low-level solver of search-based solvers. As for the high-level solvers, Conflict-Based Search (CBS)~\cite{sharon2015conflict} is a popular MAPF solver that is complete and optimal, and its variants~\cite{barer2014suboptimal, li2021eecbs} sacrifice the optimality for faster computation time. Besides, Priority-Based Search (PBS)~\cite{ma2019searching}, which solves the conflicts by priority orderings, shows prominent computational efficiency but is incomplete and not optimal.

\subsection{Lifelong MAPF}
In contrast to one-shot MAPF, lifelong MAPF solves problems that an agent may be assigned more than one task, which is always accompanied by the online scenario. In online MAPF, the tasks will be generated on the fly during execution~\cite{ma2017lifelong}. Therefore, the solution can not be planned in advance and should be evaluated by \emph{throughput} (the number of tasks finished per timestep) because the execution time and the tasks assigned to agents can be endless.
For instance, endless tasks come at any time in warehouses, and agents are assigned new tasks after they finished the previous ones. Therefore, the path solver needs to consider the solution constantly in real-time and the problems should be solved as lifelong MAPF. In addition, we should consider not only the \emph{throughput} but also the \emph{runtime} (computing time of planning) that makes sure the planning of paths can be completed on time to effectively utilize agents.

\subsection{Lifelong Solutions}
Firstly, if all the tasks are known in advance, a lifelong MAPF problem can be solved as a one-shot MAPF problem~\cite{nguyen2019generalized}. 
\fong{Otherwise, the second type of methods~\cite{wan2018lifelong, grenouilleau2019multi} replans paths for all the agents at each timestep to handle new tasks on the fly.
Nonetheless,} these two types of methods are really time-consuming in solving the whole problem offline or constantly replan at each timestep, respectively. Therefore, the third type of method~\cite{vcap2015complete} can be followed in that new paths are replanned for the agents which get new goals only\fong{, but it still encounters poor throughput without cooperative planning among all agents.}

To retain the efficiency of runtime while maximizing the throughput, RHCR~\cite{li2021lifelong} incorporated windowed MAPF~\cite{silver2005cooperative} into lifelong MAPF. The idea is to plan the entire paths to goals and solve conflicts only for the first $w$ timesteps and the $w$ represents the time horizon. In addition to the time horizon $w$, RHCR has another user-specified parameter $h$ that specifies the replanning period. Every $h$ timesteps as an episode, the new goals are assigned to the agents which have reached their goals, and the conflict-free paths in subsequent $w$ timesteps are planned for the future when the agents are following the planning results from the previous episode. After $h$ timesteps pass, agents follow these new conflict-free paths, and the paths for the next episode should be planned. However, due to the incompleteness and the lack of guarantee in the optimality, the windowed MAPF solver encounters deadlocks (i.e., agents idle at their current locations waiting for each other to pass first), which rely on a potential function to evaluate the progress of the agents and $w$ will be increased to address deadlocks~\cite{li2021lifelong}. Besides, because windowed MAPF~\cite{silver2005cooperative} only solves conflicts in the first few timesteps and replans paths constantly, it causes agents to revise their path direction or revisit locations that they had previously visited, which also happens in RHCR~\cite{li2021lifelong}.

\subsection{Highway}
In one-shot MAPF, the \emph{highway} is an add-on setting on the map to speed up the runtime, which builds a global rule of the moving direction to encourage or enforce MAPF solvers to search paths in a consistent direction.
Forbidding agents to move against the highway direction is a simple strategy to reduce computational complexity and avoid collisions. For instance, a one-shot solution~\cite{wang2008fast} used directed maps to reduce the complexity of MAPF for faster computational efficiency.
Additionally, another one-shot solution~\cite{cohen2015feasibility} increases the cost on the paths against the highway direction when calculating the heuristic values, which implicitly encourages solvers to explore the path along the highway direction.
However, these methods are proposed for one-shot MAPF where undesirable phenomena in lifelong MAPF such as deadlocks and rerouting do not occur. Even if RHCR~\cite{li2021lifelong} (a lifelong solution) used directed maps in its experimental section to reduce the planning complexity, it still lacked further discussions and experiments on how \emph{highway} affects the planning results in lifelong scenarios.

\section{Problem Definition}

The multi-agent path finding (MAPF) problem is formulated as a directed graph of the map $G=(L, E)$ and a set of agents $A=\{ a_{1},...,a_{k}\}$.
The graph represents the locations $L$ and the edges $E$ connecting pairs of these locations.
For each agent $a_i$, it starts from its start location $l^{i}_{s}$ and aims at its own goal location $l^{i}_{g}$. 
Assuming that time is discretized into timesteps, $a_i$ should decide to \emph{move} to a neighboring location $l_i \in L$ through the connected edge $e \in E$ or \emph{wait} at its current location at each timestep, and the movement will be completed within the timestep.

However, a collision may occur when two agents arrive at the same location or move through the same edge at the same time, which is referred to as a \emph{conflict}~\cite{stern2019multi}.
Therefore, the objective of MAPF is to find collision-free paths $P=\{ p_{1},...,p_{k}\}$ for all the agents to reach their goals jointly, where $p_{i} = [l^{i}_{s},...,l^{i}_{g}]$ is the path of agent $a_{i}$ as a sequence of neighboring locations.

We assume the scenario as lifelong and online MAPF. The agents start from their locations without knowing their task (i.e., moving to a goal location) in advance, and a new task will be assigned to the agent after it finishes its task (i.e., reaching its goal location). The goal of the MAPF solver is to constantly plan the path for each agent and maximize the \emph{throughput} (the number of tasks finished per timestep).

\section{Lifelong MAPF with Highways}

\subsection{Highway Definition}
In one-shot MAPF with highways~\cite{cohen2015feasibility}, \emph{highway} was represented as a subgraph $G_{H}=(L_{H}, E_{H})$ of the given graph of the map $G=(L, E)$, where $L_{H}$ contained locations involved in the \emph{highway} and $E_{H}$ represented the edges following the \emph{highway} direction.

\begin{figure}[t]
\centering
\includegraphics[width=0.7\columnwidth]{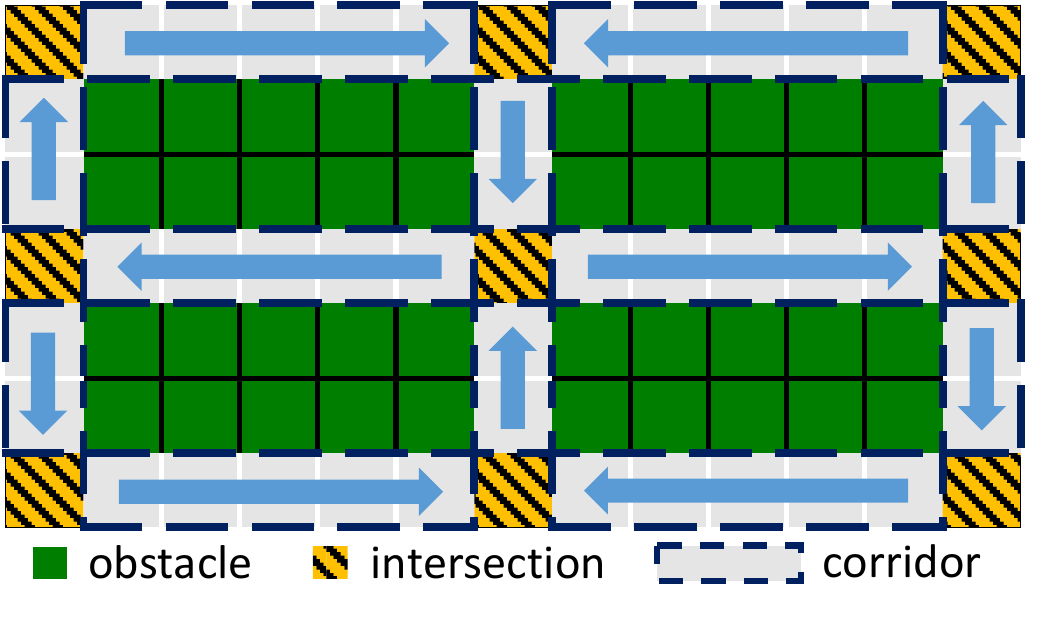}
\vspace{-4mm}
\caption{
    An example map where the arrows represent the direction of the highway in each \emph{corridor}. The \emph{intersections}, which connect \emph{corridors}, cannot be a part of any \emph{corridors}.
}
\label{fig:highway_example}
\vspace{-6mm}
\end{figure}

In this work, we focus on warehouse-like maps~\cite{stern2019multi}, the \emph{highway} settings can be defined by specifying the direction of movement for each \emph{corridor}.
A sequence of locations $K =\{l_{0},...,l_{n}\} \subseteq L$ is called a \emph{corridor} with length $n$ iff $l_{i}$ is connected to exact two locations $l_{i-1}$ and $l_{i+1}$ according to $E$ for $i=1 \rightarrow i-1$ (i.e., only one way in and one way out at each location). The $l_{0}$ and $l_{n}$ can be the same location as a \emph{loop-corridor}. For each \emph{corridor}, there exists a direction $d \in D$, which is a \emph{bool-value} representing the \emph{direction} of the \emph{corridor}, shown in Fig.~\ref{fig:highway_example}. Besides, any location as an \emph{intersection}, which connects two or more \emph{corridors}, will not belong to any \emph{corridors}. Finally, given a set of \emph{directions} $D=\{d_{0},...,d_{m}\}$ (m is the number of \emph{corridors}), the \emph{highway} can be represented as a subgraph $G_{H}=(L_{H}, E_{H}) \subseteq G$ and the edges that against the highway direction $E_{H'}= \{(l_b, l_a) | (l_a, l_b) \in E_{H}\} $.

\subsection{Lifelong MAPF Framework}
We follow RHCR~\cite{li2021lifelong} as our lifelong MAPF framework. Given user-specified time horizon $w$ and replanning period $h$ ($h \leq w$), the conflict-free paths in the first $w$ timesteps will be planned for agents every $h$ timesteps, and the new goals will be assigned to the agents which have reached their goals. As for the high-level solver, we use PBS~\cite{ma2017feasibility}, which shows prominent computational efficiency with RHCR. 

In our method, we use A* as the low-level solver with the true shortest distance heuristic. The shortest-path heuristic values, which ignore dynamic constraints (i.e., moving agents), can be calculated by shortest-path algorithms in advance. In this way, the precise values of the distances between nodes can guide the low-level solver to traverse fewer locations to the goal than the heuristic values of Manhattan distance. Besides, given a \emph{highway} setting, the calculated heuristic can influence the low-level solver to find paths following the direction of \emph{highway}, thereby resulting in fewer conflicts, \fong{and the complexity of low-level planning is also relaxed because choices to move against the highway direction are trimmed away.}


\subsection{Strict-limit Highway and Soft-limit Highway}
There are two methods to set up the highway for the low-level solver. Firstly, the movement across edges is allowed for only one direction~\cite{wang2008fast}, which we call \emph{strict-limit highway} in this work. The second one is to increase heuristic values only except following the direction of the highway~\cite{cohen2015feasibility} that encourages low-level solvers to search paths along the edges of the highway, which we call \emph{soft-limit highway} here.

\subsubsection{Strict-limit Highway}
While using the strict-limit highway, $G=(L, E)$ will be replaced with the subgraph $G_{S}=(L, E_{S})$, where $E_{S} = (E - E_{H'}) \subseteq E$. Therefore, the connectivity of the neighboring locations is restricted strictly under the strict-limit setting, which means the movement against the highway is impossible here, as shown in Fig.~\ref{fig:highway_diff}. The strict rule forces agents to follow the directions of the highway but limit the flexibility to use idle locations, because the low-level solver may no longer explore neighboring locations that avoid the highway direction.
To utilize the highway with the strict limitation in RHCR, the heuristic values based on the all-pair shortest paths ignoring moving agents should be calculated and stored for usage in advance.
\fong{The heuristic computed with $G_{S}$ is defined as: 
\begin{gather}\label{eq_h}
    H_{S}(l_s,l_g) = \min_p |p|-1
\end{gather}}
where $p = [l_s, . . . , l_g]$ is a feasible path from a start location $l_s$ to the goal location $l_g$ and $|p|$ is the length of the path. Hence, \fong{$H_{S}(l_s,l_g)$} represents the shortest distance moving from start $l_s$ to goal $l_g$.
Then, the lower-level solver follows the guidelines from the precise highway heuristic to make agents move along the direction of the highway.

\subsubsection{Soft-limit Highway}
\begin{figure}[t]
\centering
\includegraphics[width=0.65\columnwidth]{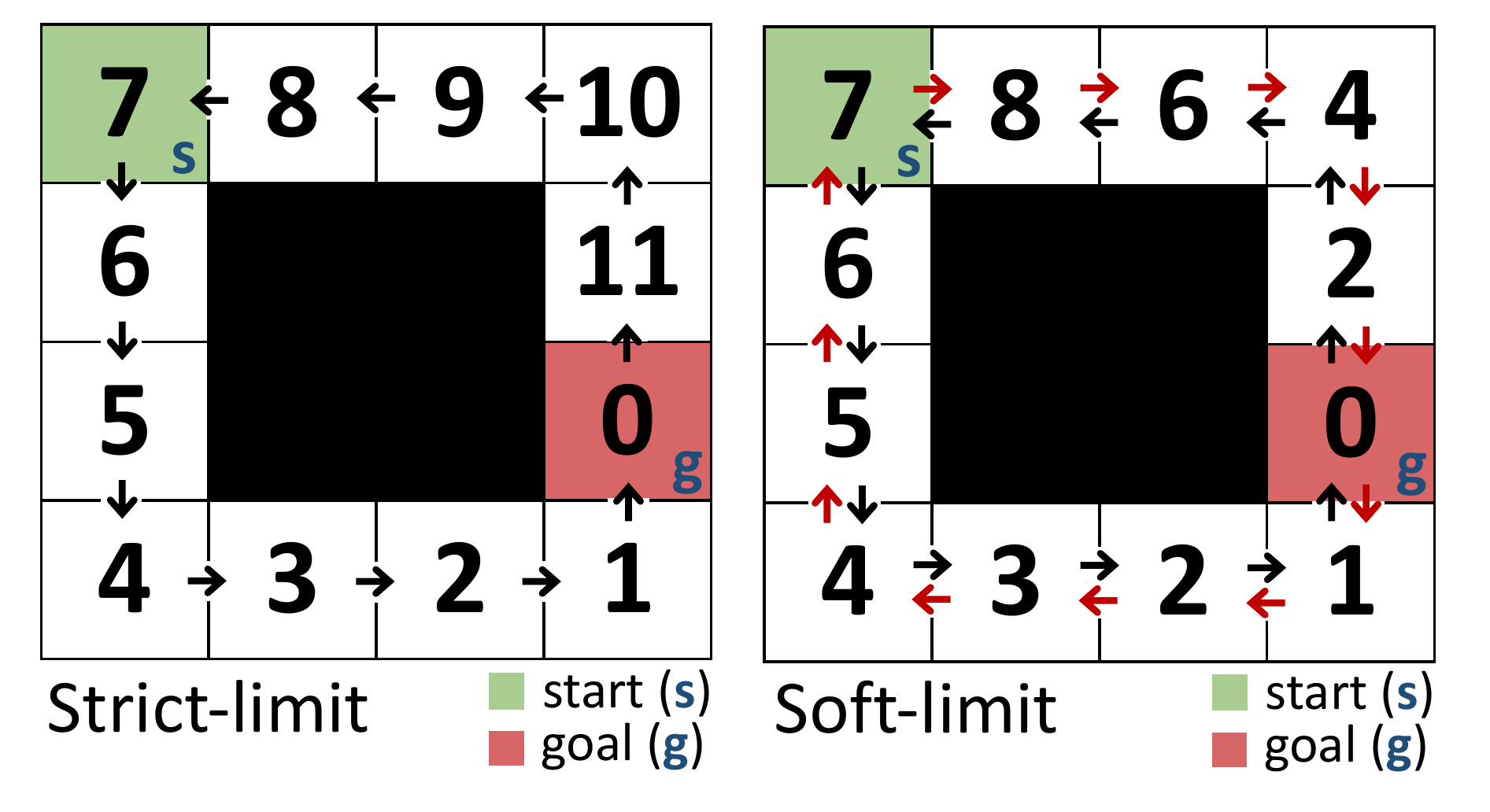}
\vspace{-1mm}
\caption{Comparison of the connectivity between \emph{strict-limit} (left) and \emph{soft-limit} (right) with $c$ = 2. The direction of \emph{highway} is counterclockwise and the black arrows represent the movement that agents can take for \emph{strict-limit}. \emph{Soft-limit} allows bidirectional movement (both black and red arrows), but \emph{strict-limit} does not. The numbers are the heuristic values from the location to the goal, which are strictly decreasing along the highway direction in \emph{strict-limit}. In \emph{soft-limit}, when the shortest path contains movement against the highway direction, the cost is increased to $c$ (i.e., 8, 6, 4, 2).}
\label{fig:highway_diff}
\vspace{-4mm}
\end{figure}

The limitation is added indirectly to the heuristic while low-level planning. There is a user-specified parameter $c$ greater than one. During the calculation of the heuristic values, the cost of crossing an edge along the direction of the highway is normally one, and the cost is increased to $c$ when moving in the opposite direction of \emph{highway},
which is defined as: 
\begin{gather}\label{eq_h}
    H(l_s,l_g,c) = \min_p\sum_{(l_i,l_{i+1})\in p}
    \begin{cases}
    c, 
        & \text{if } (l_i,l_{i+1})\in E_{H'}\\
    1,
        & \text{otherwise},
    \end{cases}
\end{gather}
where $p = [l_s, . . . , l_g]$ is a feasible path from a start location $l_s$ to the goal location $l_g$.
When $c$ is close to one, the heuristic values are close to the shortest paths without any limitation. On the contrary, if the $c$ is increasing to infinity, the heuristic values are similar to the shortest paths computed under the highway setting. The highway heuristic values encourage the agents to move in the same direction to avoid collisions. 
For instance, an example with $c=2$ is shown in Fig.~\ref{fig:highway_diff}, and the heuristic values on the path moving against the highway direction are higher (i.e., 8, 6, 4, 2).
However, because of the lack of strict limitations, the movement against the highway direction will still happen, which means there exist conflicts that should be solved. 
To use the \emph{soft-limit highway} in RHCR, the heuristic values should be calculated according to the chosen $c$ beforehand, and then the lower-level solver will follow the heuristic to find paths. 


\subsection{Highway Behaviors}
\begin{figure*}[t]
\centering
\subfigure[Without the dynamic obstacle]{\includegraphics[width=0.35\textwidth]{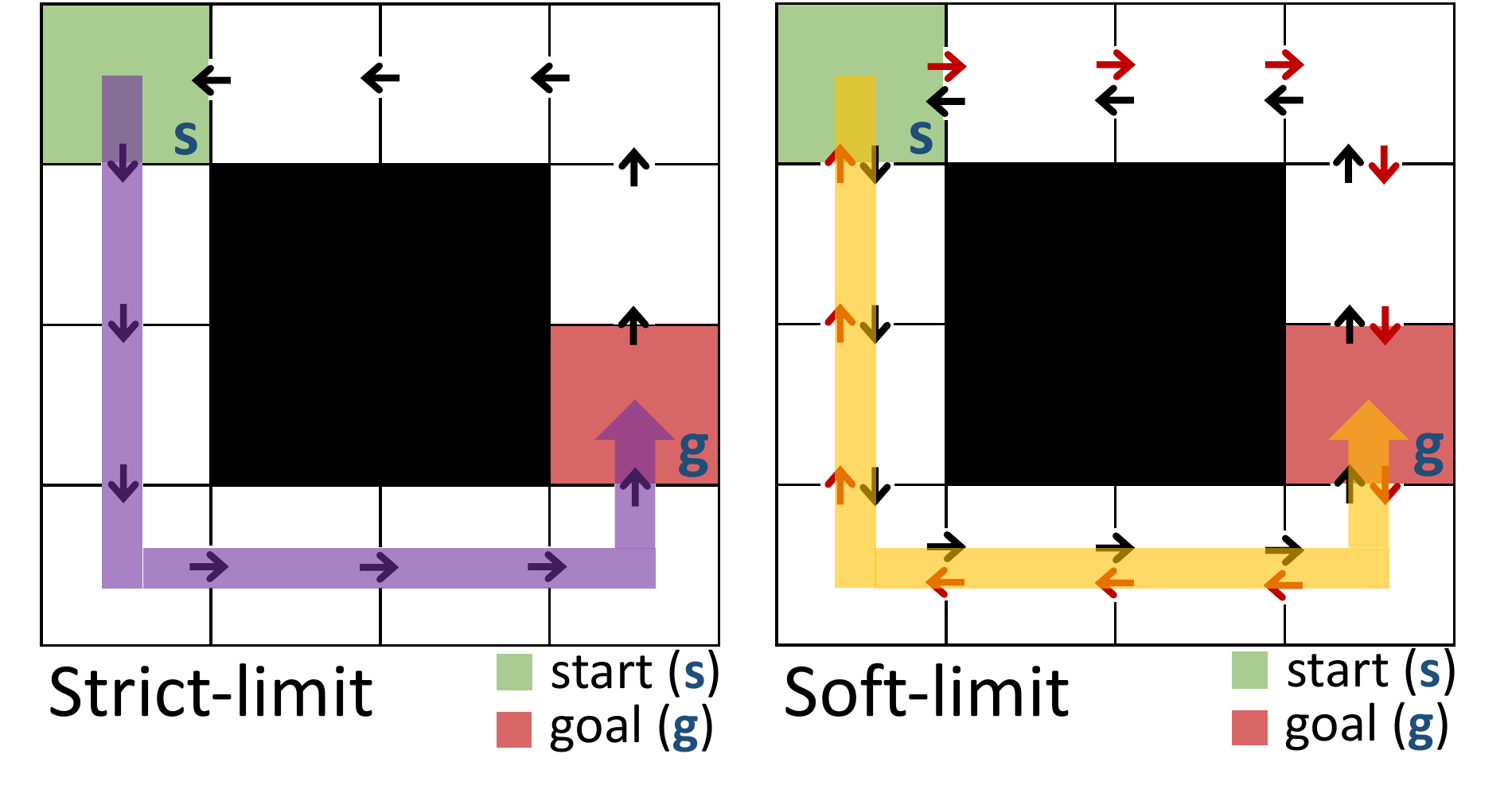}} 
\subfigure[With the dynamic obstacle]{\includegraphics[width=0.35\textwidth]{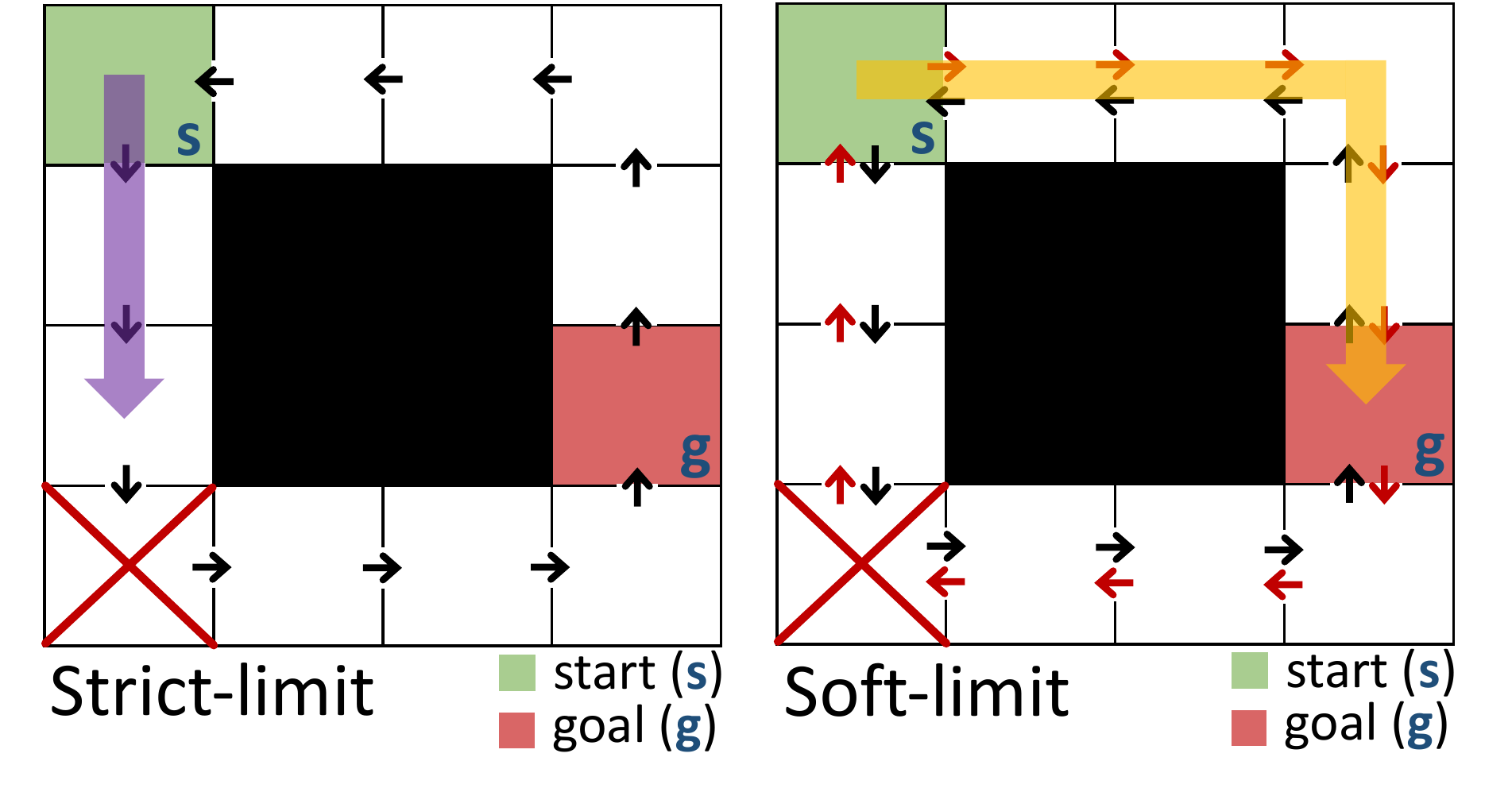}} 
\vspace{-1mm}
\caption{Different behaviors under \emph{strict-limit} and \emph{soft-limit} highway. The arrow represents the path in the next \emph{n} timesteps. When the road ahead is clear, both the agent under \emph{strict-limit} and \emph{soft-limit} highway follow the direction of the highway, shown in panel (a). When a dynamic obstacle occurs, the agent under \emph{strict-limit} highway still strictly keeps the direction of the highway and, however, the agent under \emph{soft-limit} highway chooses to reroute, as shown in panel (b).}
\label{fig:soft-strict}
\vspace{-4mm}
\end{figure*}

Under the soft-limit highway, the low-level solver is planning with the highway heuristic. Because the path against the highway direction has higher heuristic values, the neighboring locations along the highway direction have higher priority to be taken out from the open-set of A* for the lower expected costs. Hence, the direction of paths found by the low-level solver shows consistency in the direction. The consistency of paths for agents causes fewer face-to-face collisions, which decreases the nodes the high-level solver should generate to solve conflicts. When $c$ is small, the results are close to the ones without the highway. As $c$ increases, the planning results are gradually close to the ones using the shortest paths of the highway as the heuristic values. However, even if $c$ is set to infinity, the movement against the highway direction happens when the locations along the shortest path are blocked by the dynamic obstacle, which shows the flexibility for agents to choose the direction of their movement but may cause conflicts that take time to solve. Using the strict-limit highway, the consistent direction of paths of agents can be guaranteed by blocking the edges to neighboring locations during planning, as shown in Fig.~\ref{fig:soft-strict}. 

The purpose of utilizing \emph{highway} is to ensure the paths of agents can share a collaborative rule locally, which relieves the congestion by solving conflicts as well as keeps the consistency of the planning results in neighboring time horizons. Benefiting from the consistent direction of paths, the strict-limit highway has the following properties addressing the issues of deadlocks and rerouting. Additionally, we show that the soft-limit highway can exploit these properties depending on the given $c$ in our experimental section.

\subsection{Property of Avoiding Deadlocks}
\begin{figure}[h]
\centering
\subfigure[Bi-directed, $w$ = 2 and 3]{\includegraphics[width=0.3\textwidth]{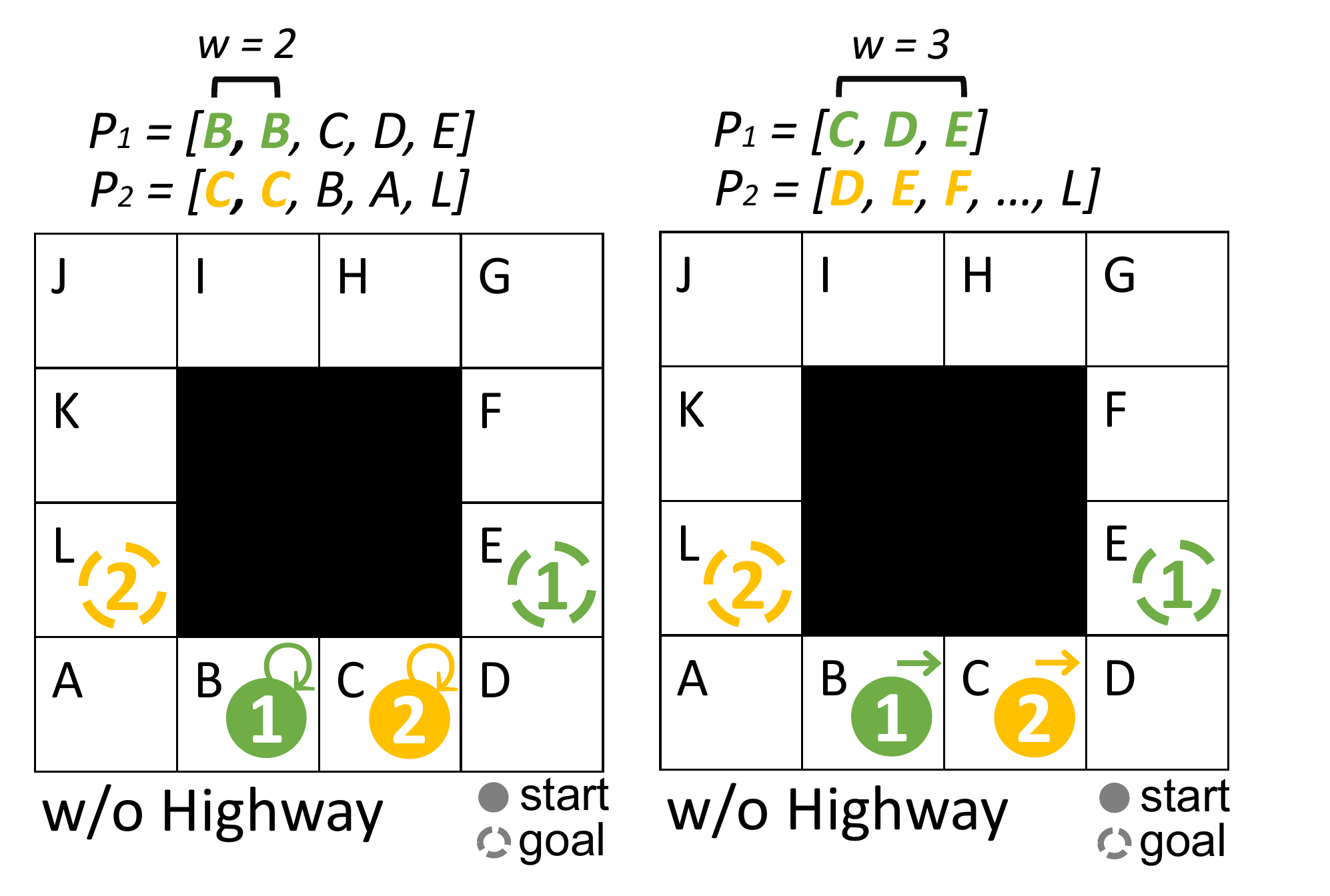}} 
\subfigure[Bi-directed, $w$ = 2]{\includegraphics[width=0.16\textwidth]{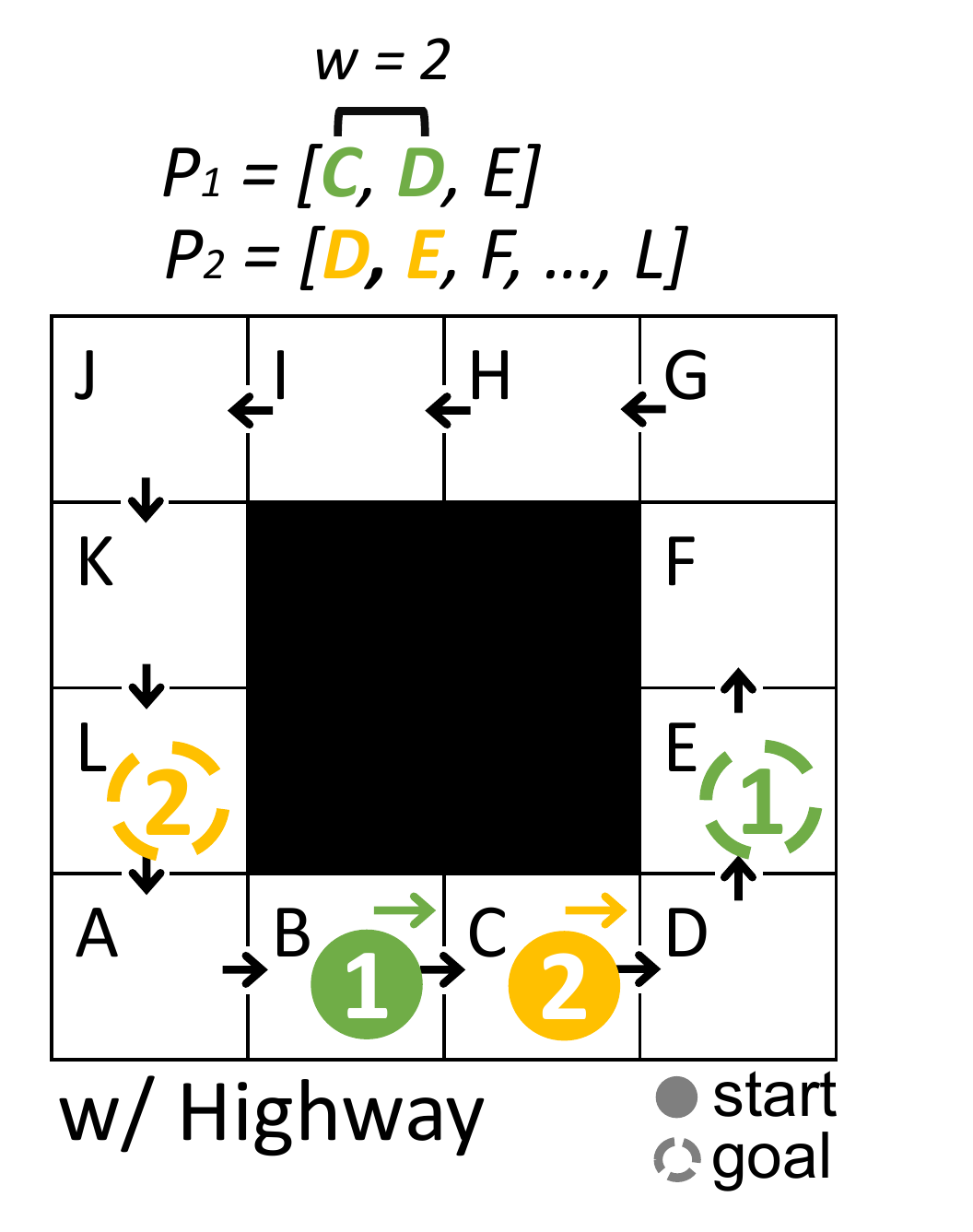}}
\vspace{-1mm}
\caption{A case of the \emph{deadlock} situation. $P_{1}, P_{2}$ represent the paths of $a_{1}, a_{2}$ found by the MAPF solver and $w$ is the \emph{time horizon}. In panel (a) without \emph{highway}, the $w$ should be set to 3 to avoid \emph{deadlock}. In panel (b), \emph{deadlock} won't happen under \emph{highway} using any $w$.}
\label{fig:deadlock}
\vspace{-4mm}
\end{figure}

\emph{Deadlock} is a phenomenon caused by the \emph{windowed MAPF} without considering the entire time horizon~\cite{li2021lifelong}. A deadlock happens when two agents move face-to-face with their goals on opposite sides. With a small $w$, neither they can go directly because of the swapping conflict nor go along the reverse direction because the high-level solver prefers the agents waiting at the original location for less ~\emph{sum-of-costs} instead of taking a longer path. For example, if time horizon ${w=2}$, replanning period ${h=2}$ and CBS is used as the high-level solver, the windowed MAPF solver returns the paths in Fig.~\ref{fig:deadlock}-(a), which lets agents stay at the location without moving for $w$ timesteps and start to move after that to the goal location. This situation causes by the limited cooperative planning of the windowed MAPF solvers, which only solves the conflicts in the first $w$ timesteps and ignores the conflicts behind that. Hence, the low-level solver is not able to consider the further locations because the \emph{sum-of-costs} is smaller while waiting at the original location. To make one agent move along the top side and the other move along the bottom side, $w$ should be set to $3$ at least to let the solver consider the further path instead of a cheating solution that waiting across the time horizon to avoid conflicts like in Fig.~\ref{fig:deadlock}-(b). Worse yet, the bigger value of $w$ should be checked to find the solution if the corridor is longer or the goals are further, which increases the runtime. 

Generally, a deadlock happens when the MAPF solver can not find feasible moving paths for agents with lower \emph{sum-of-costs} compared with paths of waiting. That is, the MAPF solver can not find the new locations for agents in the next $w$ timesteps that have a lower sum of distances to their goals. Given $A_d \subseteq A$ represents a subset of agents which may cause a deadlock. An agent $a_i \in A_d$ asks for $p_i = [l^i_c, . . . , l^i_n]$ ($|p_i|-1  = w$, where $w$ is the time horizon), which is a feasible path from its current location $l^c_i$ to the next location $l^i_n$ in the next $w$ timesteps without collisions. With the shortest path heuristic values that represent the distances to the goal locations in the strict-limit highway, a situation that a deadlock may happen can be simplified as:
\begin{gather}\label{eq_dlk1}
    \sum_{a_i \in A_d}{H_{S}(l_c^i, l_g^i)} \leq \sum_{a_i \in A_d}\min_{p_i} H_{S}(l_n^i, l_g^i)
\end{gather}
where $l_g^i$ is the goal location of $a_i$.

In \emph{strict-limit highway}, if an agent $a_i$ moves from any location $l^i_a$ to its neighboring location $l^i_b$ and $(l^i_a,l^i_b) \in E_{H}$ always, then $H_{S}(l^i_b, l^i_g)=H_{S}(l^i_a, l^i_g)-1$ if $l^i_a \neq l^i_g$, where $l^i_g$ is its goal location. Thus, Eq.~\eqref{eq_dlk1} can be rewritten as:
\begin{gather}\label{eq_dlk2}
    \sum_{a_i \in A_d}{H_{S}(l_c^i, l_g^i)} \leq \sum_{a_i \in A_d}\min_{p_i} H_{S}(l_c^i, l_g^i) - |\bar{p}_i| + 1
\end{gather}
where $\bar{p}_i$ is the set of unique locations in $p_i$, and $|\bar{p}_i|$ reflects the number of unique locations in $p_i$.

Eq.~\eqref{eq_dlk2} is not held when $\exists a_i \in A_d, |\bar{p}_i| > 1$. Namely, in the strict-limit highway, if all agents move along the edges $\in E_{H}$, deadlocks will not happen unless every agent has no available neighboring locations to move to.

\subsection{Property of Avoiding Rerouting}
\emph{Rerouting} is a phenomenon that results from the lack of long-term consideration of \emph{windowed MAPF}~\cite{silver2005cooperative}, which causes agents to revise their path direction or revisit locations that they had previously visited. The idea of the \emph{windowed MAPF} used in RHCR is to separate the lifelong MAPF problem into a sequence of subproblems by time, which is referred to as \emph{episode}. In each \emph{episode}, the low-level solver plans the entire path for each agent, and then the high-level solver solves the conflicts of the paths for the finite time horizon $w$. After planning, each agent follows the path of the first $h$ (smaller than or equal to $w$) timesteps. However, although planning the entire path to the goal, only the partial path in the first $h$ timesteps is followed by each agent. Therefore, each agent may stop at a location whose heuristic value is higher than the original location, namely waiting is a wiser choice rather than moving. Besides, in most search-based MAPF solvers, special mechanisms are followed to determine the priority orderings in high-level planning to decide which agent should go first when a conflict happens (e.g. CBS~\cite{sharon2015conflict} adds the constraint to the certain agent as a \emph{CT-node}, and PBS~\cite{ma2019searching} maintains priority ordering pairs). However, the consistency of the priority orderings of agents is not guaranteed in the subsequent high-level planning, and changes in the priority orderings may lead to an obvious difference in the planning results. 
These cases may ask agents to \emph{reroute} or even move backward, resulting in longer distances for agents to reach their goals.

We formally defined that an agent $a_i$ is \emph{rerouting} when it is assigned a path $p_i = [l_c^i,..., l_n^i]$ by the MAPF solver that makes $a_i$ move away from its goal $l_g^i$, which can be represented by the shortest path heuristic:
\begin{gather}\label{eq_rt}
   H_{S}(l_c^i, l_g^i) < H_{S}(l_n^i, l_g^i)
\end{gather}
where $l_c^i$ is the current location of $a_i$ and $l_n^i$ represents the location where $a_i$ will arrive after $h$ timesteps ($h$ is the replanning period smaller than or equal to the time horizon $w$).
In \emph{strict-limit highway}, given an agent $a_i$ and its goal location $l^i_g$, $\forall (l_a,l_b) \in E_{H}$, $H_{S}(l_a, l^i_g)=H_{S}(l_b, l^i_g)+1$ if $l_a \neq l^i_g$. Therefore, when $a_i$ moves along the edges $\in E_{H}$, Eq.~\eqref{eq_rt} is not held because $H_{S}(l_c^i, l_g^i) \geq H_{S}(l_n^i, l_g^i)$.
In summary, using \emph{strict-limit highway}, these properties guarantee that there are no \emph{deadlocks} and \emph{rerouting} as agents move across the edges that are parts of the highway.


\section{Experiments}
\begin{figure*}[ht]
    \centering
    \begin{tabular}{ccc}
        \subfigure[Throughput]{\includegraphics[width=0.28\textwidth]{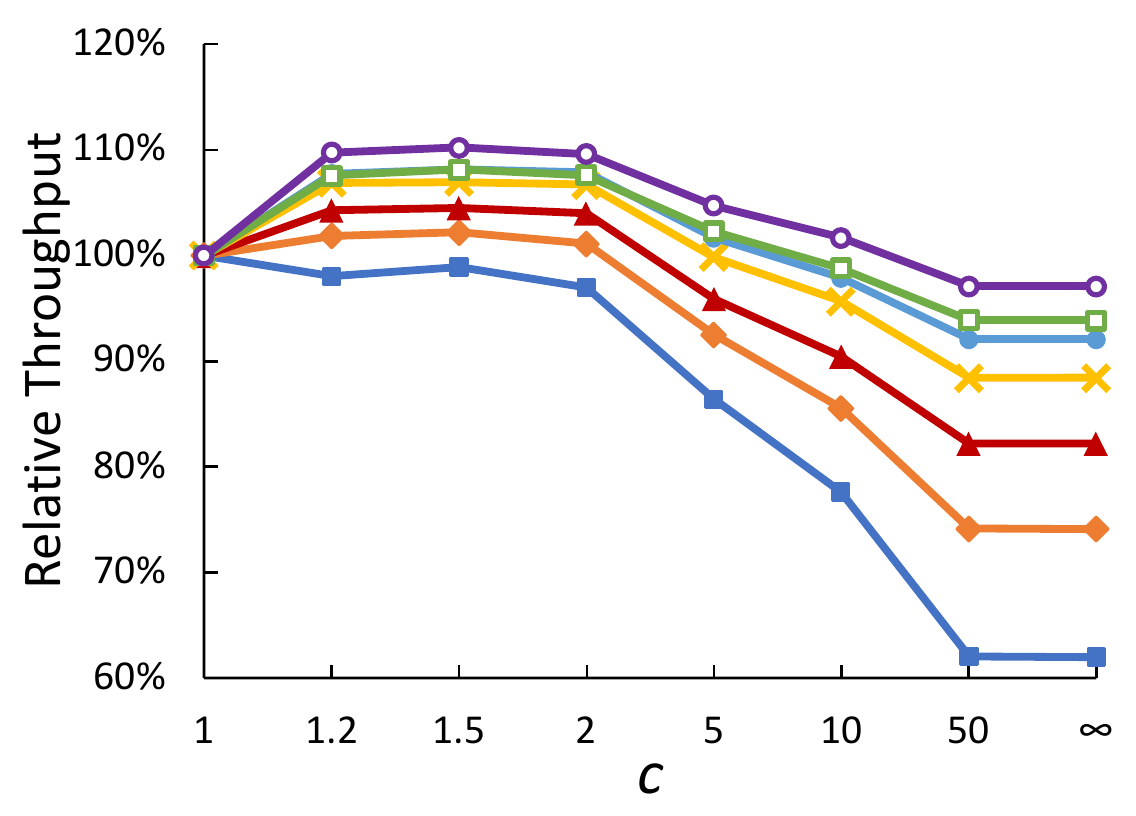}}
        &
        \subfigure[Runtime]{\includegraphics[width=0.34\textwidth]{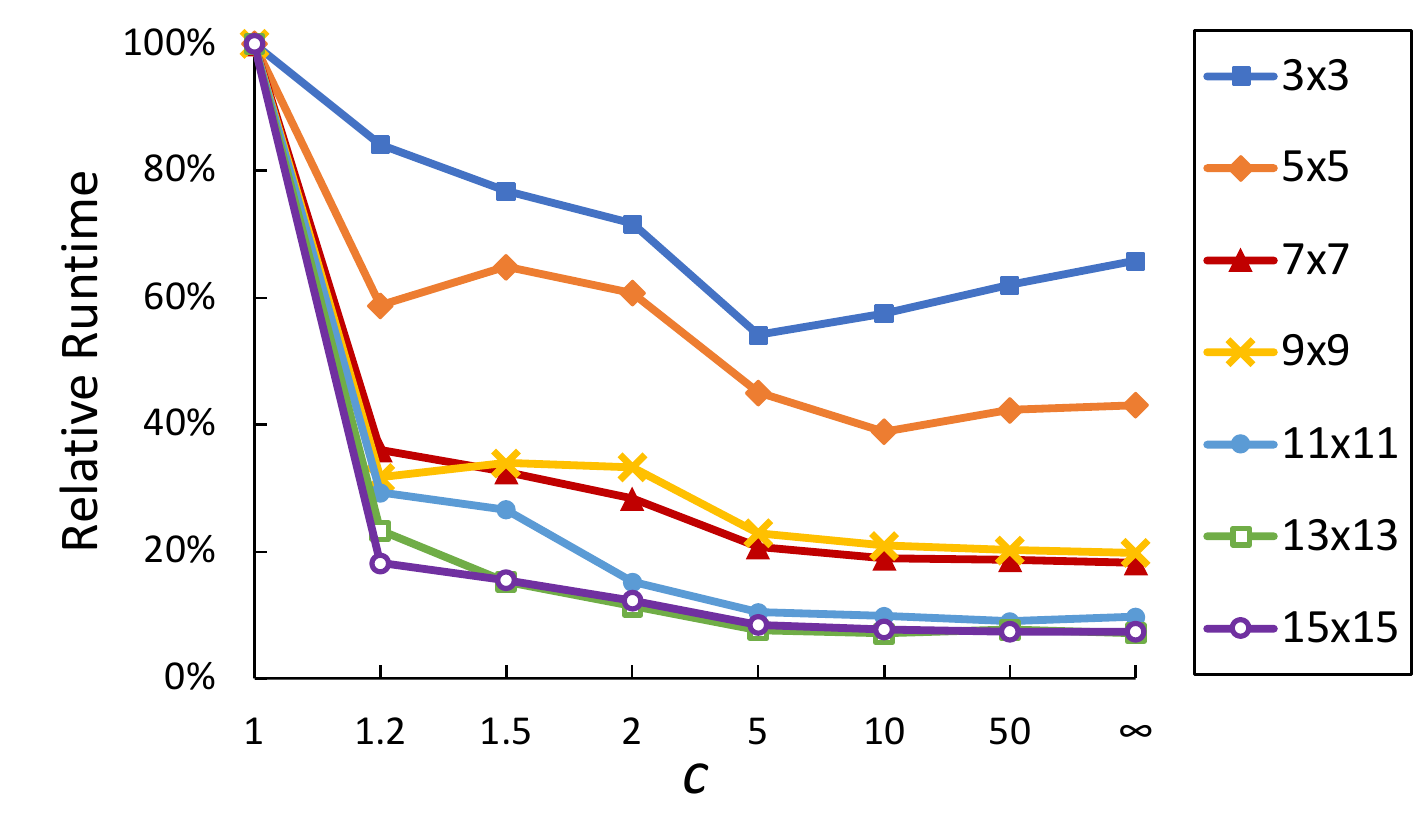}}
        &
        \subfigure[Generated nodes]{\includegraphics[width=0.29\textwidth]{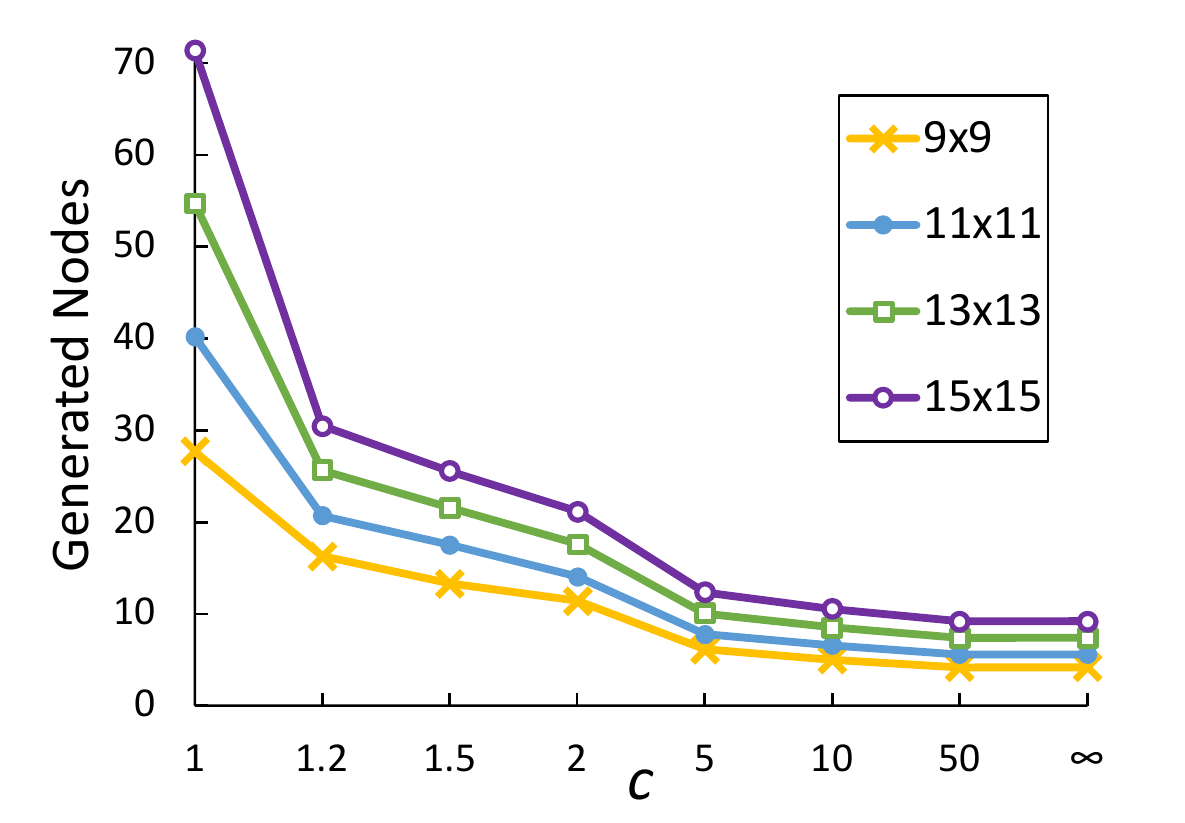}}
    \end{tabular}
    \vspace{-1mm}
    \caption{ 
        Relative ratios of throughput, runtime, and numbers of generated nodes (i.e., the number of high-level nodes generated by PBS before finding the solution)  in different map sizes. The results of different $c$ for \emph{soft-highway} are compared to the results without \emph{highway}, namely with $c=1$. As the $c$ increases, the cost of moving against the direction of the highway grows, which gradually makes agents move along the direction of the highway and results in lower throughput and faster runtime because of fewer nodes generated when planning. Moreover, the impact on throughput is reduced as the size of the map increases, which is shown from the bottom line to the top line of panel (a).
    }
    \label{fig:highway_c}
    \vspace{-4mm}
\end{figure*}

Our experiments focus on the comparison between \emph{before} and \emph{after} incorporating \emph{highway} into lifelong MAPF and the analysis of strengths and drawbacks. Firstly, we evaluated the changes in the throughput and runtime when leveraging \emph{highway}. Secondly, we analyzed the advantages of using \emph{highway} when map sizes or agent densities grow larger.

\subsection{Environment}
\label{sec:environment}
We implement RHCR with $w=5$, $h=5$ using PBS as the lifelong MAPF solver with a standard location-time A* as the low-level solver in Python and ran all experiments on Intel Core i9-9980XE with 16 GB memory. Before each round of experiments, the start locations and goal locations of agents are selected randomly from the different empty locations on the map. When an agent arrives at its goal location, a location that is not the goal location of any other agents will be assigned. Besides, we randomly initialize 100 episodes for each experiment and simulate 100 iterations of planning in each episode. Each iteration has its time limit of 60 seconds, and the episode will be labeled as a \emph{fail case} if any iteration times out. All metrics shown are calculated as averages which exclude the fail cases.

Generally, \emph{throughput} and \emph{runtime} are the two main metrics for lifelong MAPF. For evaluating that \emph{highway} can effectively deal with existing phenomena of deadlocks and rerouting, \emph{moving timesteps}, \emph{idle timesteps}, and \emph{rerouting rate} are recorded during the experiments. \emph{Rerouting rate} is the percentage of the agents who reroute in an iteration. Besides, \emph{moving timesteps} reflects the number of timesteps that an agent moves during a task and \emph{idle timesteps} is the number of timesteps that an agent stays at the same location without moving during a task. For observing how increasing \emph{c} influences agents, \emph{highway avoidance rate} shows the chance that an agent moves against the highway direction. Similar to~\cite{sharon2015conflict} and~\cite{cohen2015feasibility}, \emph{generated nodes} is used to explain the reason why our method has lower \emph{runtime}, which is the number of high-level nodes generated by the MAPF solver before finding the solution during one planning.

Given that the heuristic values are calculated from the shortest paths (e.g. \emph{strict-limit} highway and \emph{soft-limit} highway with $c=1$ or $\infty$), a trick mentioned in the previous work~\cite{silver2005cooperative} can be applied to simplify planning. Namely, in windowed MAPF, instead of planning the entire path to the goal, we can directly ignore the planning after $w$ timesteps, which yields the same result but saves the time to plan entire paths. We include this trick in Table~\ref{tab-map-size} for comparison.

\subsection{Fulfillment Warehouse}
\begin{figure}[t]
\centering
\subfigure[Warehouse map~\cite{stern2019multi}]{\includegraphics[width=0.65\columnwidth]{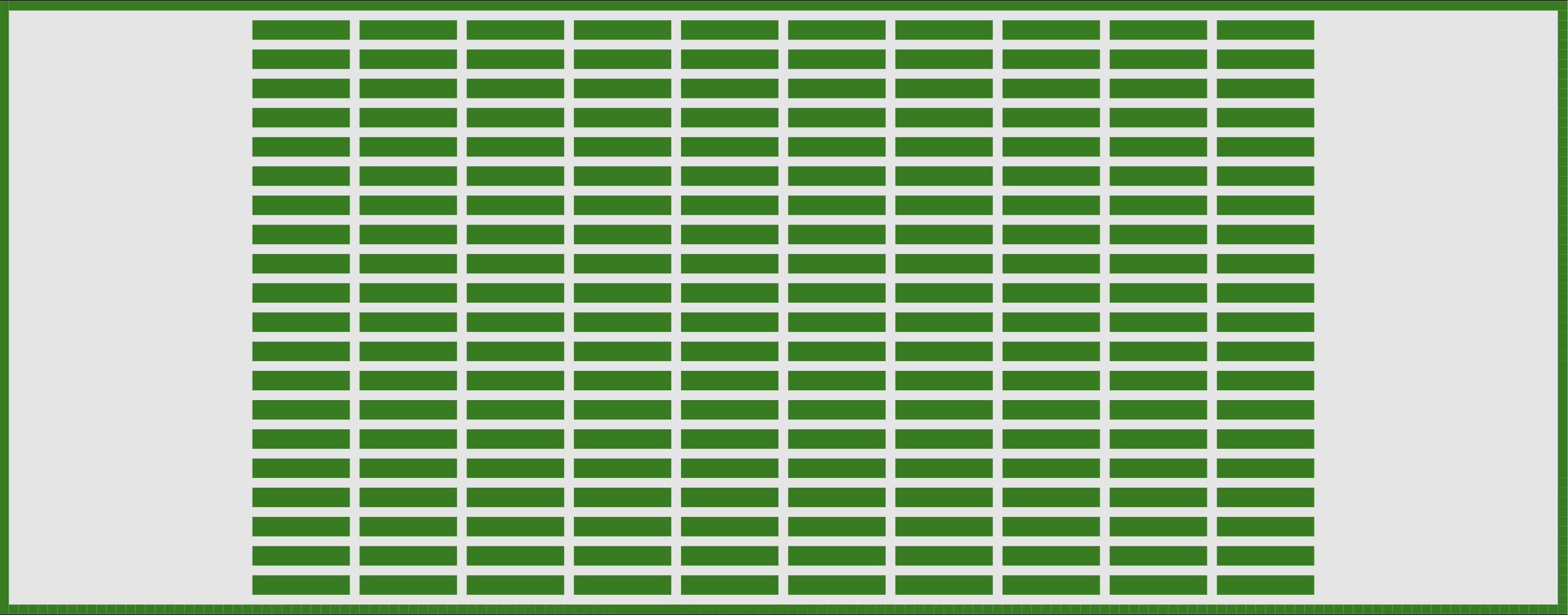}}
\subfigure[Warehouse map with directions of highway]{\includegraphics[width=0.7\columnwidth]{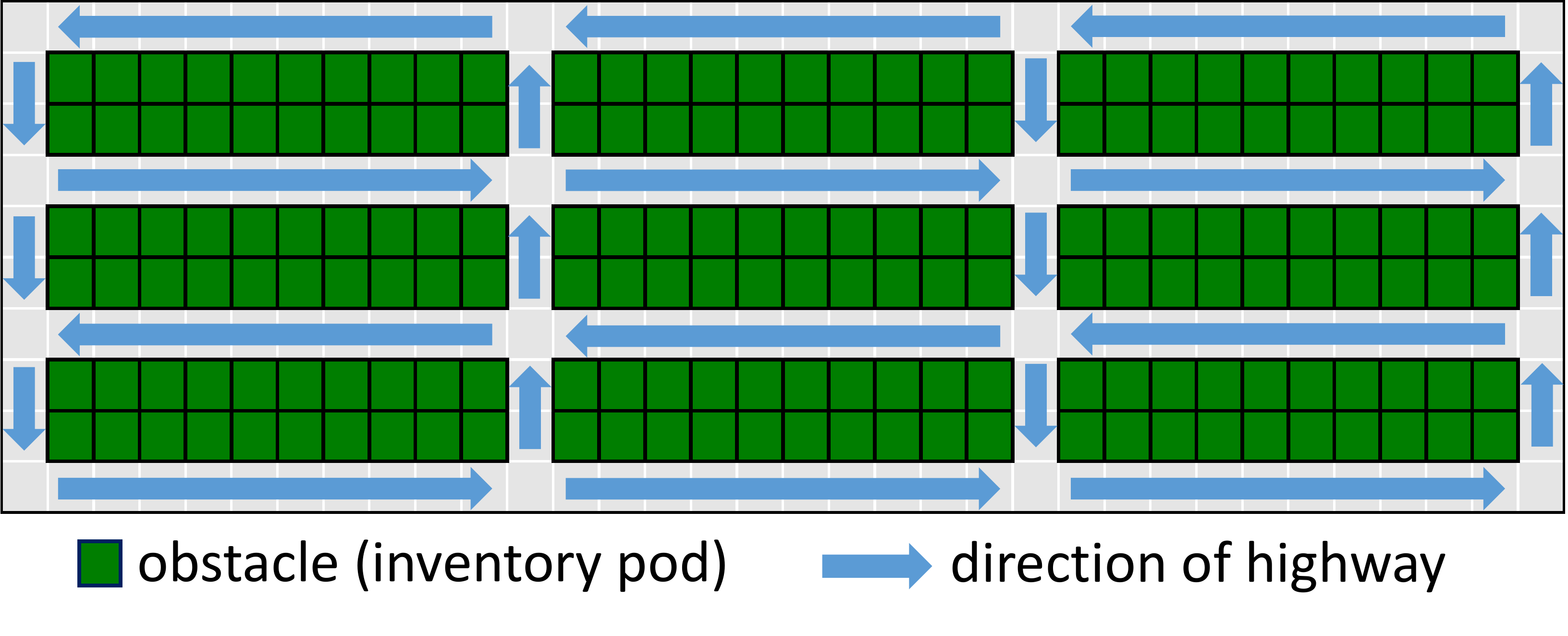}}
\vspace{-1mm}
\caption{Panel (a) shows the warehouse map in~\cite{stern2019multi} with 10$\times$20 blocks of inventory pods. Each rectangle indicates one block. Panel (b) shows the warehouse map with 3$\times$3 blocks of inventory pods with the direction (arrow) of the highway. Each block (green rectangle) in both maps consists of 10$\times$2 inventory pods (green squares).}
\label{fig:warehouse-highway-3x3}
\vspace{-4mm}
\end{figure}

In autonomous warehouses, a set of inventory pods are placed closely
into a rectangle block, and the blocks with corridors in between are placed into a grid, like in Fig.~\ref{fig:warehouse-highway-3x3}. In our experiments, we follow the settings of the \emph{warehouse map} in Fig.~\ref{fig:warehouse-highway-3x3}-(a)~\cite{stern2019multi}: (1) each block contains 10$\times$2 pods, and (2) the corridors are single-rowed. 
For simplicity, we test maps with N$\times$N blocks, where N is an odd number (i.e., 3, 5, 7, etc.), and inventory pods are considered obstacles for the agents. For instance, the smallest 3$\times$3 version of the maps is shown in Fig.~\ref{fig:warehouse-highway-3x3}-(b). 
In addition, the percentage of obstacles over the entire map indicates how crowded the map is.
Our smallest map (Fig.~\ref{fig:warehouse-highway-3x3}-(b)) has more than 50\% of obstacles, and the percentage of obstacles gradually increases as the map grows with the larger N. The high obstacle density and neighboring corridors are suitable for testing \emph{highway}.

\subsection{From No Highway to Highway}
The parameter $c$ in \emph{soft-limit highway} determines how much the map is influenced by the highway directions. Therefore, we evaluate the influence of \emph{highway} through the map in Fig.~\ref{fig:warehouse-highway-3x3}-(b) with the increasing $c$ and fixed 5\% density of agents (i.e., the ratio of agents to empty locations of the map). Fig.~\ref{fig:highway_c} shows the relative throughput, runtime, and generated nodes with the $c \in \{1, 1.2, 1.5, 2, 5, 10, 50, \infty\}$, which are compared to the result of $c=1$.

When the map is small, the value of throughput drops when the value of $c$ rises in Fig.~\ref{fig:highway_c}-(a), though the runtime decreases slightly in Fig.~\ref{fig:highway_c}-(b). However, in the case of the larger map, the drop in throughput is much slower, and the drop in runtime is significant as $c$ increases. Surprisingly, the relative throughput when $c=1.5$ even surpasses the throughput without \emph{highway}. Furthermore, a trend shows that the gap in the throughput between \emph{no-highway} ($c=1$) and \emph{highway} ($c=\infty$) narrows as the map size grows larger.

The speedup on the runtime mainly benefits from the more efficient planning under \emph{highway}, because the global rule of the direction should be followed, and thus planning results are more consistent, which causes fewer conflicts before the solution is found. Referring to Fig.~\ref{fig:highway_c}, if a higher $c$ is used, fewer nodes should be generated to find a solution, and the benefit is even more pronounced in larger maps.

In Table~\ref{tab-highway-c}, the results verified that using \emph{highway} resulted in fewer average idle timesteps (timesteps that an agent stays at the same point during a task), which reflects the consistent direction of paths and fewer deadlocks, and the average idle timesteps reduce more when a larger map. Also, through \emph{c} increases, the chance that an agent moves against the highway direction and the rerouting problem decreases, shown as the highway avoidance rate (the chance of an agent moving against the highway direction) and the rerouting rate (the percentage of the rerouting agents in an iteration of planning) in Table~\ref{tab-highway-c}. Besides, following the highway causes agents to have more moving timesteps (timesteps that an agent needs to arrive at its goal). Nevertheless, using \emph{highway} in a larger map causes fewer extra moving timesteps, which explains another reason that a larger map can benefit more while using \emph{highway}. For instance, in Table~\ref{tab-highway-c}, using \emph{highway} ($c=50$) in the 3x3 map causes 69.1\% extra moving timesteps while using \emph{highway} ($c=50$) in the 7x7 map only causes 23.9\% extra moving timesteps.

\begin{table}[t]
    \footnotesize
    \centering
    \caption{Average idle timesteps, moving timesteps, highway avoidance rate, and rerouting rate with different map sizes, \emph{c}, and the same density of agents (5\%) in the \emph{soft-limit highway} setting.}
    \begin{threeparttable}
    \begin{tabular}{l|l|ccc}
        \toprule
        & \multirow{2}{*}{c} & & \scriptsize \multirow{-1.2}{*}{Map Size} \footnotesize & \\
        & & \multirow{1.1}{*}{3$\times$3} & \multirow{1.1}{*}{5$\times$5} & \multirow{1.1}{*}{7$\times$7} \\
        \midrule
        & 1 & 2.49 & 2.66 & 2.83 \\
        Idle & 2 & 2.42 (-2.5\%) & 2.55 (-4.1\%) & 2.61 (-7.9\%) \\
        Timesteps & 5 & 2.32 (-6.8\%) & 2.38 (-10.5\%) & 2.42 (-14.7\%) \\
        &  50 & 2.22 (-10.8\%) & 2.29 (-13.8\%) & 2.33 (-17.9\%) \\
        \midrule
        & 1 & 17.65 & 27.76 & 37.69 \\
        Moving & 2 & 18.27 (3.5\%) & 27.49 (-0.9\%) & 36.42 (-3.4\%) \\
        Timesteps & 5 & 20.90 (18.4\%) & 30.61 (10.3\%) & 39.96 (6.0\%) \\
        & 50 & 29.84 (69.1\%) & 38.45 (38.5\%) & 46.72 (23.9\%) \\
        \midrule
        Highway & 1 & 43.2\% / 3.8\% & 44.3\% / 4.5\% & 44.9\% / 4.7\% \\
        Avoidance / & 2 & 32.3\% / 2.5\% & 28.3\% / 1.9\% & 25.5\% / 1.5\% \\
        Rerouting & 5 & 18.7\% / 0.7\% & 14.9\% / 0.4\% & 12.5\% / 0.3\% \\
        Rate & 50 & 2.2\% / 0.0\%  & 2.0\% / 0.0\%  & 1.8\% / 0.0\% \\
        \bottomrule
    \end{tabular}
    \begin{tablenotes}[para,flushleft]
    For comparing how much idle timesteps decrease and moving timesteps increase after using \emph{highway} (c $\neq$ 1), percentages of difference when c $\neq$ 1 compared to \emph{no-highway} $(c=1)$ are represented in parenthesis.
    \end{tablenotes}
    \end{threeparttable}
    \label{tab-highway-c}
    \vspace{-6mm}
\end{table}

\begin{table*}[t]
    \footnotesize
    \centering
    \caption{Throughput and runtime with different map sizes and the same density of agents (5\%).}
    \begin{threeparttable}
    \begin{tabular}{l|l|cccccccc}
        \toprule
        & Types $\backslash$ Map Size& 3$\times$3 &  5$\times$5 & 7$\times$7 & 9$\times$9 & 11$\times$11 & 13$\times$13 & \fong{15$\times$15} \\
        \midrule
        \multirow{3}{*}{\shortstack[l]{Throughput \\ (tasks/step)}} & Strict-limit & 0.23 (-41.4\%) & 0.45 (-28.7\%) & 0.69 (-20.5\%) & 0.93 (-14.1\%) & 1.17 (-10.6\%) & 1.43 (-8.3\%) & 1.68 (-3.8\%) \\
        & Soft-limit & 0.24 (-38.0\%) & 0.47 (-25.9\%) & 0.71 (-17.8\%) & 0.96 (-11.5\%) & 1.21 (-7.9\%) & 1.46 (-6.2\%) & 1.71 (-2.2\%) \\
        &  w/o Highway & 0.39 & 0.63 & 0.87 & 1.09 [20] & 1.31 [47] & 1.56 [77] & 1.75 [95] \\
        \midrule
        \multirow{4}{*}{\shortstack[l]{Runtime \\ (seconds)}} & Strict-limit (P) & 0.002 (6.7) & 0.010 (13.2) & 0.032 (20.4) & 0.069 (32.1) & 0.130 (30.4) & 0.284 (34.0) & 0.553 (39.8) \\
        & Strict-limit & 0.007 (1.8) & 0.034 (3.9) & 0.097 (6.7) & 0.241 (9.1) & 0.339  (11.7) & 0.681 (14.2) & 1.323 (16.6) \\
        & Soft-limit & 0.008 (1.5) & 0.041 (3.2) & 0.111 (5.8) & 0.321 (6.9) & 0.439 (9.0) & 0.870 (11.1) & 2.016 (10.9) \\
        & w/o Highway & 0.012  & 0.133  & 0.646  & 2.208  & 3.955  & 9.638 & 21.983 \\
        \bottomrule
    \end{tabular}
    \begin{tablenotes}[para,flushleft]
    Each is an average calculated from test cases excluding fail cases (i.e, time out). The results using \emph{strict-limit highway}, \emph{soft-limit highway} ($c=\infty$) and the baseline result without \emph{highway} are shown respectively, and the result using the partial planning trick (mentioned in Section~\ref{sec:environment}) to speed up is marked as "(P)". Percentages of throughput decay using the highway compared to \emph{no-highway} are represented in parenthesis, and numbers of fail cases are marked in square brackets in the top column if any (out of 100 test cases). The ratios of speedups using the highway compared to \emph{no-highway} are represented in parenthesis in the bottom column.
    \end{tablenotes}
    \end{threeparttable}
    \label{tab-map-size}
    \vspace{-4mm}
\end{table*}

\subsection{Scaling up}
To ensure the runtime is efficient enough to assign the paths on time without idling the agents, the question is whether the advantages of \emph{highway} are kept after the map of a warehouse scales up. Referring to the results mentioned previously, the answer is clearly yes. Subsequently, we discuss and quantify the benefits of \emph{highway} as the map size scales up and the agent density grows larger.

\subsubsection{Map Size} 
To scale up the size of the map, the block is kept the same as in Fig.~\ref{fig:warehouse-highway-3x3}-(b), which contains 10$\times$2 pods in each block, and corridors are still single-rowed. Then, the number of blocks are increasing from 3$\times$3,  5$\times$5, 7$\times$7 blocks ... to 15$\times$15 blocks, and the density of agents is fixed at 5\%. 
The results using \emph{strict-limit highway}, \emph{soft-limit highway} ($c=\infty$), and the baseline without \emph{highway} are shown in Table~\ref{tab-map-size}. 
When the map grows larger, the gap of throughputs between \emph{highway} and the baseline becomes smaller, and the speedup gained from using \emph{highway} keeps increasing. The main reason for the decrease in the throughput gap is that, as the map gets larger, the number of extra moves taken for the highway becomes insignificant relative to the total number of moves. Furthermore, in contrast to the setting using \emph{highway}, the fail cases appear on larger maps without \emph{highway}.

\begin{figure}[t]
\centering
\includegraphics[width=0.82\columnwidth]{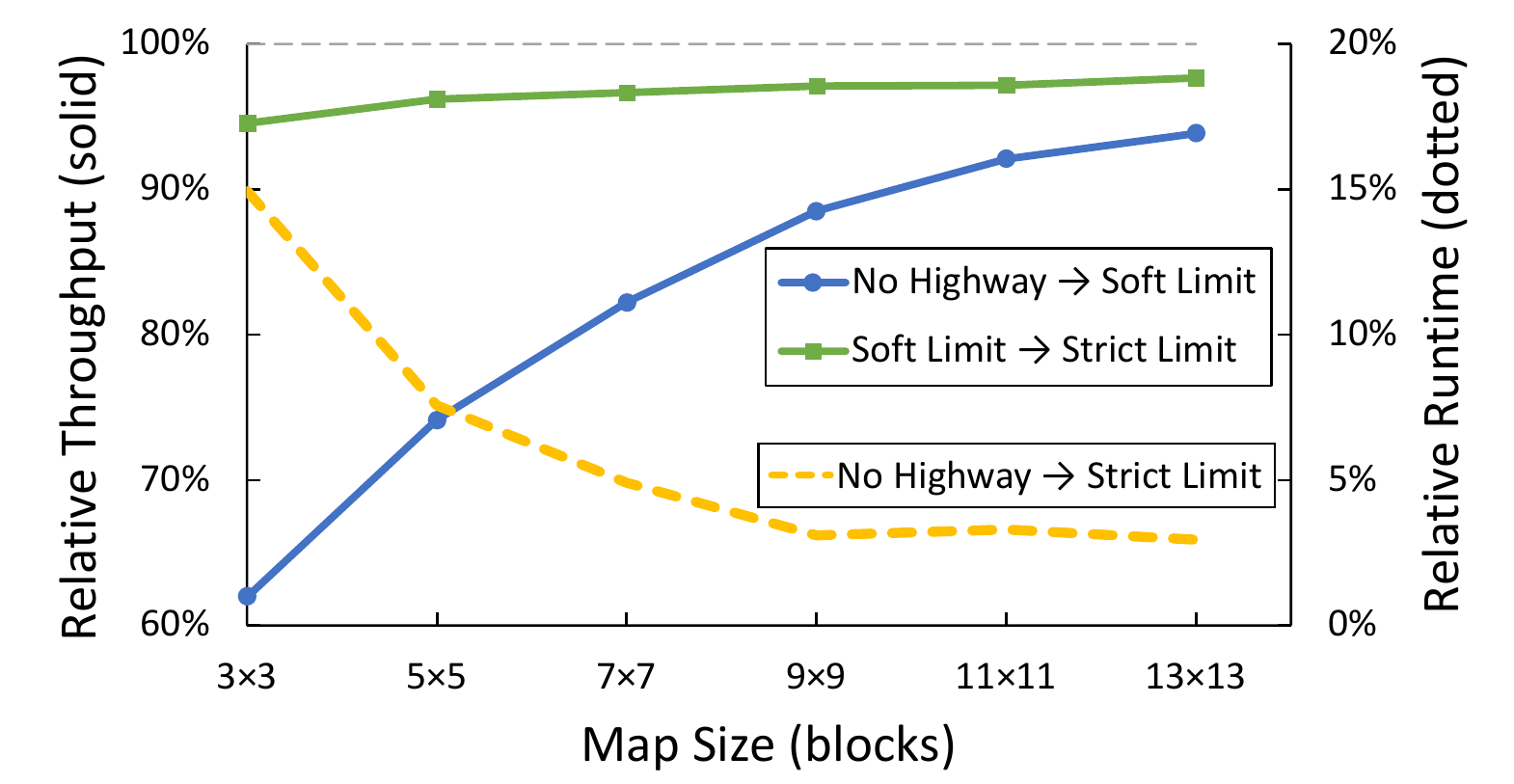}
\vspace{-2mm}
\caption{Relative throughput of transferring "\emph{no-highway} to \emph{soft-limit highway}" and "\emph{soft-limit highway} to \emph{strict-limit highway}". The dotted line represents the relative runtime of replacing "\emph{no-highway} with \emph{strict-limit highway}", showing the reduction of runtime. As the
map enlarges, the relative throughputs of these two settings approach 100\% (solid lines) and the runtime keeps decreasing (yellow dotted line).}
\label{fig:diff_map_size}
\vspace{-2mm}
\end{figure}

Besides, the gap between throughputs of \emph{strict-highway} and \emph{soft-highway} is also decreasing in Fig.~\ref{fig:diff_map_size}. 
Therefore, it becomes more cost-effective to convert \emph{soft-highway} to \emph{strict-highway} for faster runtime (see Table~\ref{tab-map-size}).

\subsubsection{Number of Agents} 
\begin{figure}[t]
\centering
\subfigure[Throughput]{\includegraphics[width=0.27\textwidth]{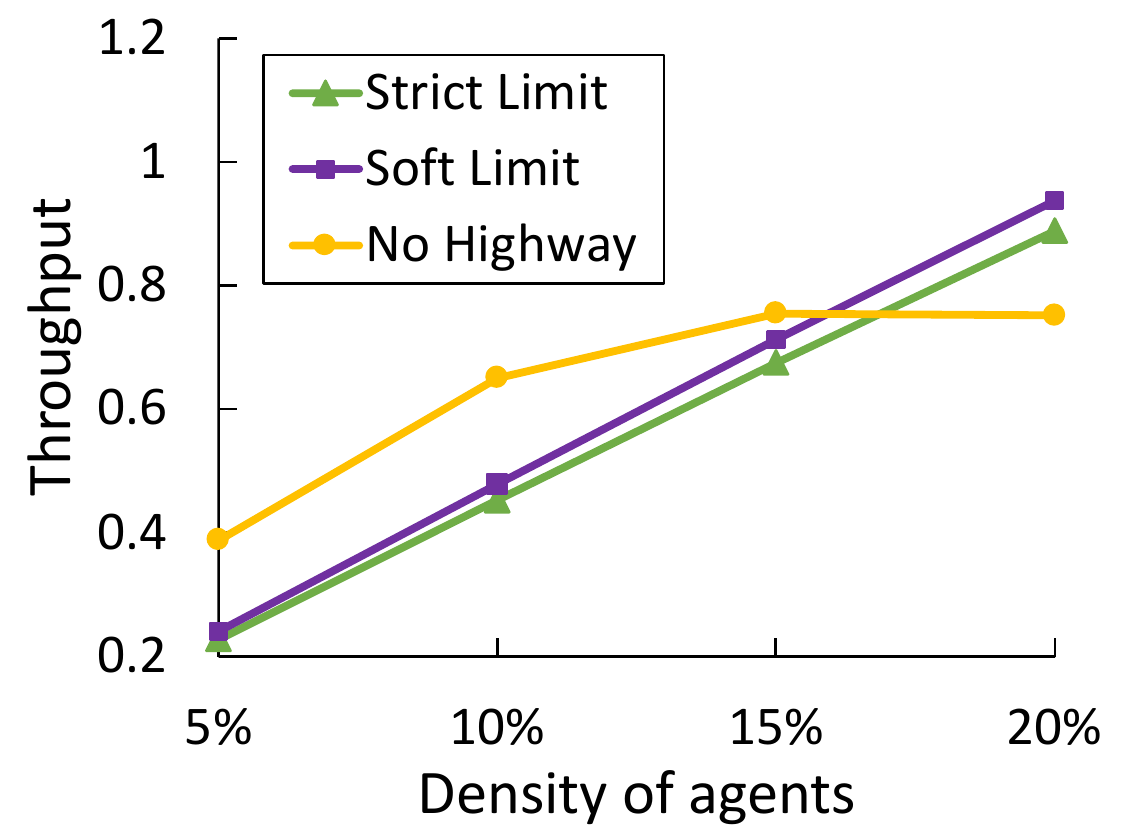}} 
\subfigure[\fong{Rerouting rate and idle steps}]{\includegraphics[width=0.24\textwidth]{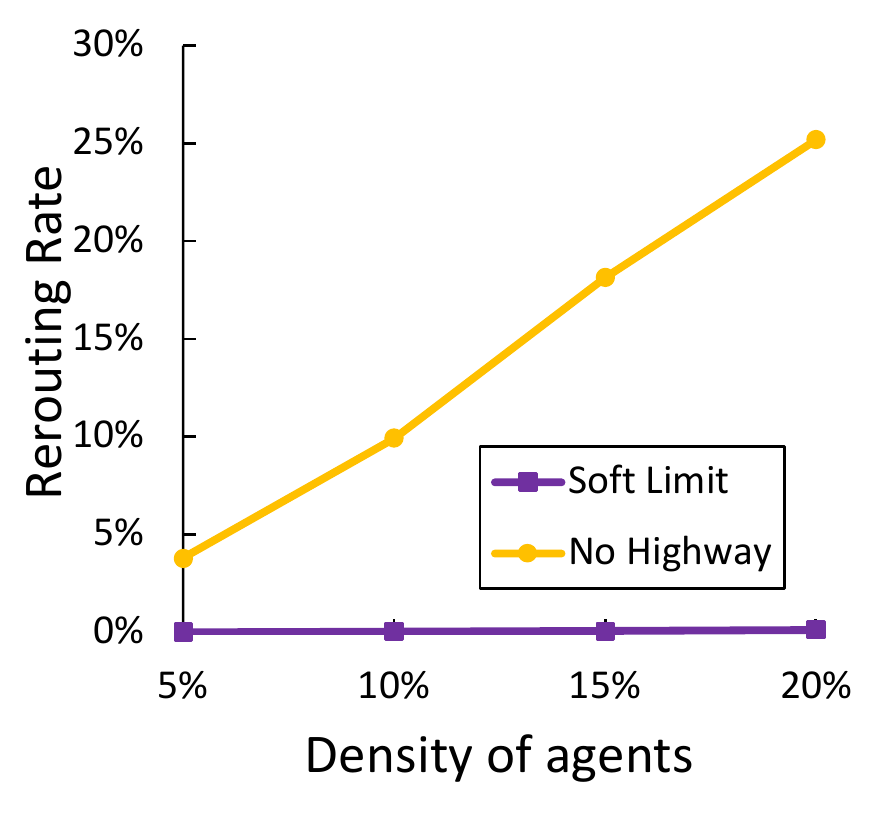}
\includegraphics[width=0.24\textwidth]{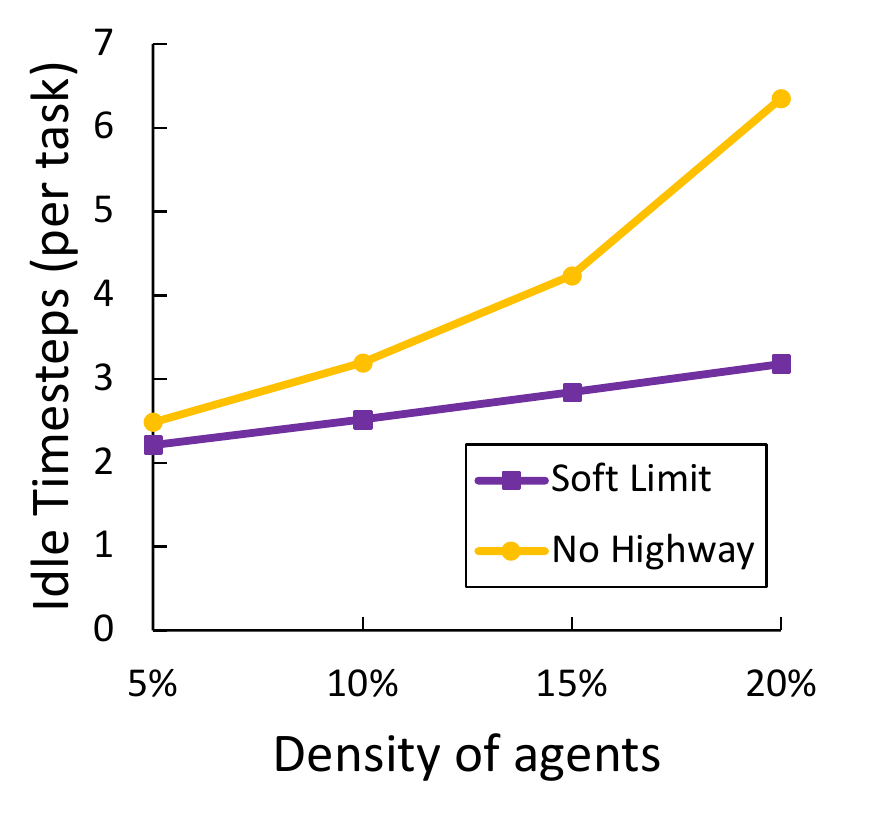}} 
\vspace{-2mm}
\caption{Comparison of throughput, rerouting rate, and the average number of idle timesteps between \emph{highway} and \emph{no-highway} in different densities of agents. When the agent density is high, using \emph{highway} even gains higher throughput than \emph{no-highway}, shown in panel (a).
As shown in panel (b), this is because the phenomena of rerouting (solid lines) and idle agents (dotted lines) are severe.}
\label{fig:diff_highway_type}
\vspace{-4mm}
\end{figure}

To test the influence of different agent densities on the map, the density of agents is increasing from 5\% to 20\% with the 3$\times$3 map in Fig.~\ref{fig:warehouse-highway-3x3}-(b). 
Referring to the result in Fig.~\ref{fig:diff_highway_type}, the throughput of using \emph{highway} increases constantly. However, the trend of the throughput without using \emph{highway} starts to decline when the density increases. Therefore, the rerouting rate and average idle timesteps are evaluated. Under the setting of \emph{highway}, \emph{rerouting} rarely happens and the fewer average idle timesteps potentially indicate that \emph{deaklocks} and the phenomena of waiting for each other are significantly reduced because of the consistent direction of paths for agents.

\section{Conclusion}
In this work, we studied the \emph{highway} idea (previously proposed for one-shot MAPF) in the lifelong MAPF scenario for the trade-off between runtime and throughput.
Furthermore, we discussed the properties of combining \emph{highway} with the lifelong MAPF framework that minimizes the existing problems of deadlocks and rerouting. 
Finally, we evaluated \emph{highway} through a sequence of experiments.
According to the experimental results, using \emph{highway} can significantly speed up the runtime, and the decay of throughput gradually diminishes when the map size or agent density grows larger.





\bibliographystyle{IEEEtran}

\end{document}